\def\year{2022}\relax
\documentclass[letterpaper]{article} % DO NOT CHANGE THIS
\usepackage{aaai22}  % DO NOT CHANGE THIS
\usepackage{times}  % DO NOT CHANGE THIS
\usepackage{helvet}  % DO NOT CHANGE THIS
\usepackage{courier}  % DO NOT CHANGE THIS
\usepackage[hyphens]{url}  % DO NOT CHANGE THIS
\usepackage{graphicx} % DO NOT CHANGE THIS
\usepackage{natbib}  % DO NOT CHANGE THIS AND DO NOT ADD ANY OPTIONS TO IT
\usepackage{caption} % DO NOT CHANGE THIS AND DO NOT ADD ANY OPTIONS TO IT
\usepackage{subcaption}
\DeclareCaptionStyle{ruled}{labelfont=normalfont,labelsep=colon,strut=off} % DO NOT CHANGE THIS
\graphicspath{{plots/}}
\usepackage{algorithm2e}

\usepackage{amsmath,amssymb,mathtools,booktabs,placeins}
\newcommand{\norm}[1]{\left\lVert #1 \right\rVert}
\newcommand{\remove}[1]{}

\title{Federated Dynamic Sparse Training: \\ Computing Less, Communicating Less, Yet Learning Better}
\author{
  Sameer Bibikar, Haris Vikalo, Zhangyang Wang, Xiaohan Chen\thanks{Xiaohan Chen is the corresponding author, and one of the two student authors who made major contributions in this work.}
}
\affiliations{
Department of Electrical and Computer Engineering, The University of Texas at Austin\\
\{bibikar,hvikalo,atlaswang,xiaohan.chen\}@utexas.edu
}

\begin{document}

\maketitle

\begin{abstract}
Federated learning (FL) enables distribution of machine learning workloads from the cloud to resource-limited edge devices. Unfortunately, current deep networks remain not only too compute-heavy for inference and training on edge devices, but also too large for communicating updates over bandwidth-constrained networks. In this paper, we develop, implement, and experimentally validate a novel FL framework termed \emph{Federated Dynamic Sparse Training} (\textbf{FedDST}) by which complex neural networks can be deployed and trained with substantially improved efficiency in both \textbf{on-device computation} and \textbf{in-network communication}. At the core of FedDST is a dynamic process that extracts and trains sparse sub-networks from the target full network. With this scheme, ``two birds are killed with one stone:'' instead of full models, each client performs efficient training of its own sparse networks, and only sparse networks are transmitted between devices and the cloud. Furthermore, our results reveal that the dynamic sparsity during FL training more flexibly accommodates local heterogeneity in FL agents than the fixed, shared sparse masks. Moreover, dynamic sparsity naturally introduces an ``in-time self-ensembling effect'' into the training dynamics, and improves the FL performance \textbf{even over dense training}.  In a realistic and challenging non i.i.d. FL setting, FedDST consistently outperforms competing algorithms in our experiments: for instance, at any fixed upload data cap on non-iid CIFAR-10, it gains an impressive accuracy advantage of $10\%$ over FedAvgM when given the same upload data cap; 
the accuracy gap remains $3\%$ even when FedAvgM is given $2\times$ the upload data cap, further demonstrating efficacy of FedDST. Code is available at: \mbox{\url{https://github.com/bibikar/feddst}.}
\end{abstract}

\section{Introduction}
Driven by the desire to protect the privacy of personal data and enable machine learning (ML) at the edge, Federated Learning (FL)~\cite{fedavg, problems} has recently emerged as the \textit{de facto} paradigm enabling distributed ML on a large number of client devices. In a FL system, a central cloud server mediates information transfer within a network of clients which must keep their local data private. Classical methods~\cite{fedavg} in FL involve a number of synchronous \emph{rounds}; in each round, FL runs several local training epochs on a subset of devices using only data available locally on each device.  After this local training, the clients' model updates, rather than the local data, are sent over to the central server which then aggregates them all to update a global model. 

In FL systems, heavy computational workloads are dispatched from the cloud to resource-limited edge devices. To enable usage at the edge, a FL system must optimize both device-level local training efficiency and in-network communication efficiency. Unfortunately, current ML models are typically too complex for inference at edge devices, not to mention training. Besides model compactness, communication efficiency between the cloud and devices is also desirable. Client devices, such as mobile phones, often have severe upload bandwidth limitations due to asymmetric internet connections, so reducing the upload cost of federated learning algorithms is of paramount importance. Much prior work in communication-efficient FL has focused on structured and sketched sparsity in FL updates~\cite{konecny2017}, optimal client sampling~\cite{ribero2020communicationefficient}, and other classical methods.

With the goal of producing lightweight models for inference on edge devices, significant efforts have been made towards optimizing sparse neural networks (NNs) \cite{gale2019state,chen2020lottery,chen2021lottery,ma2021sanity}. These methods significantly reduce inference latency, but heavily impact compute and memory resources needed for training. The lottery ticket hypothesis \cite{frankle2018the} demonstrated that dense NNs contain sparse matching subnetworks that are capable of training in isolation to full accuracy \cite{pmlr-v119-frankle20a}. More works show sparsity can emerge at initialization \cite{lee2018snip,Wang2020Picking} or can be exploited in dynamic forms during training \cite{rigl}.

The overarching goal of this paper is to develop, implement, and experimentally validate a novel FL framework termed \textit{Federated Dynamic Sparse Training} (\textbf{FedDST}), by which complex NNs can be deployed and trained with substantially improved efficiency of both on-device computation and in-network communication. At the core of FedDST is a judiciously designed federated approach to dynamic sparse training \cite{rigl}. FedDST transmits clients' highly sparse matching subnetworks instead of full models, and allows each client to plug in efficient sparse distributed training - thus ``killing two birds with one stone.'' More importantly, we discover that dynamic sparsity during FL training accommodates local heterogeneity in FL more robustly than state-of-the-art algorithms. Dynamic sparsity itself leads to an in-time self-ensembling effect \cite{intimeoverparameterization} and improves the FL performance even over dense training counterparts, which echoes observations in standalone training \cite{intimeoverparameterization}. We summarize our contributions as follows: 
\begin{itemize}
    \item For the first time, we introduce dynamic sparse training to federated learning and thus seamlessly integrate sparse NNs and FL paradigms. Our framework, named \textit{Federated Dynamic Sparse Training} (\textbf{FedDST}), leverages sparsity as the unifying tool to save both communication and local training costs.
    
    \item By using flexible aggregation methods, we deployed FedDST on top of FedAvg~\cite{fedavg} with no additional transmission overhead from clients. As a general design principle, our method is readily extendable to other FL frameworks such as FedProx \cite{fedprox}. Furthermore, the notion of dynamic sparsity is found to accommodate local heterogeneity, as well as create the bonus effect of in-time self-ensembling, that improve FL performance even over the dense baseline.
    
    \item Extensive experiments demonstrate that FedDST dramatically improves communication efficiency on difficult problems with pathologically non-iid data distributions. Even in these non-iid settings, FedDST provides a $3\%$ accuracy improvement over FedAvgM \cite{fedavgm} on CIFAR-10, while requiring only half the upload bandwidth. We also provide extensive ablation studies showing robustness of FedDST to reasonable variations of its parameters. These results suggest sparse training as the future ``go-to'' option for FL.
\end{itemize}

\section{Related Work}
\subsection{Federated Learning}
In \emph{federated learning}~\cite{problems}, a set of clients $j \in [N]$
collaborate to learn one or multiple models with the involvement
of a central coordinating server. Each client has a small set of
training data $\mathcal D_j$ for local training, but to preserve user privacy,
clients do not share their local training data. In this work,
we attempt to learn a global model $\theta$, and aim to minimize the global
loss

\begin{equation}
\min_{\theta} \sum_{j \in [N]} \sum_{z \in \mathcal D_j} \ell(\theta; z).
\end{equation}

In the FedAvg~\cite{fedavg, fedavgm} family of algorithms, training proceeds
in \emph{communication rounds}. To start each round $i$, the server selects
a set of clients $C_i$ and sends the current server model parameters $\theta^i$
to the clients. Each client $j \in C_i$ performs $E$ epochs of
training on the received model using its local training set to produce
parameters $\theta_j^i$, which it uploads to the server; in FedAvg, client-local training is performed via SGD. The server then updates the global model with a weighted average of the sampled clients' parameters.
Reddi et al\@. \cite{reddi2021adaptive} argue that the updates
produced by clients can be interpreted and used as pseudo-gradients.
They thus generalize FedAvg into the FedOpt framework, which allows
``plugging in'' different client and server optimizers.

Real-world FL settings present many challenges due to non-iid data distributions, client heterogeneity, and limited compute, memory, and bandwidth at the edge \cite{problems,hong2021federated}.
Heterogeneity in the compute capabilities of clients causes the so-called \emph{straggler problem},
in which certain clients take ``too long'' to form model updates and the server must proceed without them.
Hsu et al\@~\cite{fedavgm}, in FedAvgM, demonstrate that adding a momentum term to the
client optimizer consistently improves performance in non-iid settings.
Li et al\@.~\cite{fedprox} propose FedProx which adds a \emph{proximal penalty} to FedAvg and allows stragglers to submit partial updates.

For communication-efficient FL,
Konecny et al\@.~\cite{konecny2017} distinguish between \emph{sketched updates},
in which model updates are compressed during communication but not
during local training; and \emph{structured updates}, in which
local training is performed directly on a compressed representation.
Relatively little prior art discusses pruning or weight readjustment
throughout the entire FL process.

\subsection{Network Pruning}
\emph{Model pruning} aims to select a sparse subnetwork from a larger NN by removing connections. Traditionally, pruning methods start from a
highly overparameterized trained model, remove connections,
and fine-tune the pruned model.
Common goals of pruning include saving compute, memory, communication, or other resources. Many selection criteria are possible, including
weight magnitude \cite{han2015deep}, optimal brain damage
\cite{optimaldamage,optimalsurgery,layer_optimalsurgery},
zero activations \cite{apoz}, and Taylor expansions \cite{molchanov17}.

Other recent studies have proposed a variety of algorithms to
perform ``single-shot'' pruning at initialization.
Lee et al\@.~\cite{lee2018snip} select connections at initialization
by sampling a minibatch and sorting connections by their
sensitivity.
Wang et al\@.~\cite{Wang2020Picking} similarly sample a minibatch
at initialization but instead attempt to preserve gradient flow
after pruning.

\subsection{Dynamic Sparse Training}
\emph{Dynamic sparse training} (DST) 
shifts the selected subnetwork regularly throughout the training process,
maintaining a constant number of parameters throughout.
The seminal work~\cite{Mocanu2018} proposed the SET algorithm which iteratively prunes the smallest
magnitude weights and grows random connections. SET also
maintains a particular distribution of model density by layer,
following the Erd\H{o}s-R\'{e}nyi random graph topology
which scales the density of a layer with the number of
input and output connections.
In RigL~\cite{rigl}, the authors initialize the sparsity mask randomly and perform layer-wise magnitude pruning and gradient-magnitude weight growth.
As in \cite{Mocanu2018}, they follow particular layer-wise
sparsity distributions and introduce the ERK
sparsity distribution for convolutional layers, thus scaling their
density by both number of connections and kernel size.
~\citet{intimeoverparameterization} demonstrate the benefits that DST gains from parameter exploration; specifically,
by exploring a number of possible sparse networks, DST is able to
effectively perform ``temporal self-ensembling,'' allowing for
performance advantages even over dense networks \cite{liu2021deep,liu2021sparse}.

\subsection{Pruning in Federated Learning}
\label{federatedpruning}
To our knowledge, there are only two works that address pruning
throughout the FL process.
PruneFL~\cite{prunefl} relies on an initial mask
selected at a particular client, followed by a FedAvg-like algorithm
that performs mask readjustment every $\Delta R$ rounds.
Training is then performed via sparse matrix operations.
On mask readjustment rounds, clients are required to upload full dense gradients which the server uses to form the aggregate gradient $g$. When selecting a mask,
the indices $j$ corresponding to prunable weights are sorted by $g_j^2 / t_j$,
where $t_j$ is an estimate of the time cost of retaining connection $j$ in the network. The estimates $t_j$ are determined experimentally by measuring
the time cost of one round of FL with various sparsities.

LotteryFL~\cite{lotteryfl} takes inspiration from LG-FedAvg \cite{lg_fedavg}
and allows clients to maintain local representations by selecting a local subset of the global network. It can also be described as an extension of FedAvg in which each client $c$ maintains a separate mask $m_c$. At each round $r$, selected clients evaluate their subnetworks
$\theta^r \odot m_c^r$ using local validation sets. If the validation
accuracy exceeds a predefined threshold and the client's current sparsity
$\norm{m_c^r}_0$ is less than the target sparsity, then magnitude pruning is
performed to produce a new mask $m_c^{r+1}$ and the corresponding weights
are reset to their initial values.

\textbf{Comparing FedDST to prior pruning works in FL}. 
First, unlike LotteryFL, which produces a system of sparse models that only perform well on local datasets, FedDST 
produces one global sparse model, dynamic over time, that performs well everywhere. FedDST performs mask readjustments on both clients and server but these are relatively low-overhead operations (layer-wise magnitude pruning and gradient-magnitude growth). Moreover, FedDST transmits neither dense models nor gradients, and does not train a dense model even at the very beginning: this is in sharp contrast to LotteryFL and makes FedDST significantly lighter.

Second, unlike PruneFL, FedDST is designed for a challenging, realistic non-iid FL setting. For this reason, FedDST's aggregation and dynamic sparse training, which allow mask readjustments at the client, provide resilience to non-iid data in a way that PruneFL's adaptive pruning criteria cannot achieve. In particular, PruneFL's clients do not readjust masks after the first round; they instead transmit gradients to the server on certain rounds, and the server uses gradient magnitudes and layer times to decide the mask for the next round. Since data heterogeneity means that gradient magnitudes cannot be directly compared between clients, gradient aggregation is inherently unstable. FedDST provides stable updates by deciding on the mask at the server using only weight magnitudes and mask ``votes'' submitted by the clients. Our experiments demonstrate advantage of FedDST vs. PruneFL in terms of the generalization power. 

Thanks to the fixed sparsity budget throughout training, FedDST updates require very little network bandwidth, even in the worst case. Though PruneFL also transmits sparse updates for most of the rounds, it transmits full dense gradients to the server every few rounds to facilitate mask readjustments. LotteryFL requires clients to transmit dense models unless their accuracy meets a certain threshold, so dense transmissions happen even more frequently.

\section{Methodology}

%%%%%%%%%%%%%%%%%%%%%%%
% Algorithm 1: FedDST %
%%%%%%%%%%%%%%%%%%%%%%%

\begin{algorithm}[!t]
\SetKwFor{ForParallel}{for}{do in parallel}{endfor}
\KwIn{Clients $[N]$ with local datasets $\mathcal D_i$ \\
Sparsities by layer $\mathbb S = \{s^1, \dots, s^L\}$ \\
Update schedule $\Delta R, R_\mathit{end}, \alpha^r$}
 Initialize server model $(\theta^1, m^1)$ at sparsity $\norm{m^1}_0 = S$\;
 \For{each round $r \in [R]$}{
 Sample clients $C_r \subset [N]$\;
 Transmit the server model $(\theta^r, m^r)$ to all clients $c \in C_r$\;
 \ForParallel{each client $c \in C_r$}{
 Receive $(\theta_c^r, m_c^r) \gets (\theta^r, m^r)$ from the server\;
 \For{each epoch $e \in [E]$}{
 Sample a minibatch $\mathcal B$ from $\mathcal D_c$\;
 Perform one step of local training of local sparse network $\theta_c^r \odot m_c^r$ on $\mathcal B$\;
 \If{$r \mod \Delta R = 0$ and $e = E_p$ and $r < R_\mathit{end}$}{
 Perform layer-wise magnitude pruning $(\theta_c^r, m_c^r) \gets \mathrm{prune}(\theta_c^{r}; \mathbb S_{1-\alpha^r})$ to attain sparsity distribution $\mathbb S_{1-\alpha^r}$\;
 Perform layer-wise gradient magnitude growth $(\theta_c^r, m_c^r) \gets \mathrm{grow}(\theta_c^r, g_c^r; \mathbb S)$ to attain sparsity distribution $\mathbb S$\;
 }
 }
 Transmit the new $(\theta_c^{'r}, m_c^{'r})$ to the server
 (do not transmit the mask if it has not changed)\;
 }
 Receive the updated client-local networks and masks $(\theta_c^{'r}, m_c^{'r})$
 from clients $c \in C_r$\;
 Aggregate networks $\theta^{r+1/2} \gets A(\{\theta_c^{r+1}, m_c^{r+1}\}_{c\in C_r})$\;
 Perform layer-wise magnitude pruning $(\theta^{r+1}, m^{r+1}) \gets \mathrm{prune}(\theta^{r+1/2}; \mathbb S)$ to attain sparsity distribution $\mathbb S$\;
 }
 \label{feddst_algo}
 \caption{Overview of the proposed Federated Dynamic Sparse Training (FedDST).}
\end{algorithm}

FedDST provides a fully federated approach to dynamic sparse
training of NNs. In this method, we aim to learn a single
model that provides good accuracy to all clients, while also consuming
minimal compute, memory, and communication resources.
Our method is designed to perform well even in pathologically non-iid
settings.

\subsection{FedDST: Overview of the General Framework}
\label{feddst_overview}
We begin on the server by initializing a server network $\theta^1$
and a sparse mask $m^1$, following the layer-wise sparsity distribution
described in \cite{rigl}. At each round $r$, the server samples clients
$C_r$. The server network and mask are sent to clients $c \in C_r$.
Each client performs $E$ epochs of local training. After the 
$E_p$-th epoch of local training, clients perform a \emph{mask 
readjustment}, which reallocates $\alpha_r$ of the model mass to
different connections. Readjustment is only performed on certain rounds,
and the frequency of readjustment is specified by $\Delta R$. 

The selected clients upload their new sparse network and mask (if needed) to the
server, and the server aggregates the received information to produce
new global parameters and mask $(\theta^{r+1}, m^{r+1})$.
Server-side aggregation methods are discussed later in this section.

The client-side mask readjustment procedure is familiar and takes
inspiration from RigL \cite{rigl}. The goal of the mask readjustment
is to reallocate $\alpha_r$ of the model mass in order to seek
a more effective subnetwork. We first prune the network to an
even higher sparsity $S + (1-S)\alpha_r$, while maintaining the same
\emph{distribution} of weights. Then, we regrow the same number of
weights that were pruned, via gradients $g_c^r$, returning to the same original sparsity.
Because different clients in a round may produce different masks,
the server has to aggregate multiple sparse networks that may
have explored the mask space in completely different directions.
As in \cite{rigl, dettmers2019sparse}, we use a \emph{cosine decay} update schedule
for $\alpha_r$,

\begin{equation}
    \alpha_r = \frac{\alpha}{2} \left( 1 + \cos \left(
    \frac{(r-1) \pi}{R_\mathit{end}}
    \right)
    \right).
\end{equation}

On the first round, we have $\alpha_1 = \alpha$, and the proportion
of weights reallocated each time decays to $0$ at round $R_\mathit{end}$.
The parameter $\alpha$ controls the tradeoff between exploration
of the mask space and agreement between clients on mask decisions.
Larger values of $\alpha$ encourage moving more quickly around the
mask space, whereas smaller values encourage agreement and
incremental adjustment of the mask. We explore the effect of $\alpha$
 in the experiments.

\paragraph{Server Aggregation with Robustness to Heterogeneity}
\label{serveragg}
Because the server does not directly receive gradients from clients,
it must decide on the mask for the next round using only the parameters
and masks received from the clients.
From this, we define the
\emph{sparse weighted average}:
\begin{equation}\label{agg_nomem}
    \theta^{r+1/2}[i] \leftarrow \frac{
        \sum_{c \in C_r} n_c \theta_c^{'r}[i] m_c^{'r}[i]
    }{
        \sum_{c \in C_r} n_c m_c^{'r}[i]
    }.
\end{equation}

This method takes inspiration from the weighted averages used in
FedAvg \cite{fedavg}, in which the effect of a particular client
on a particular parameter is weighted by the dataset size at that client.
However, the sparse weighted average
also ignores parameter values from clients that did not provide
any value. In particular, if a client has pruned out a weight,
that client is ignored for the purposes of aggregation of that weight. Especially in pathologically non-iid FL settings, we find FedDST to benefit greatly from this mask aggregation method. Magnitudes of stochastic gradients cannot be directly compared between clients
because data distribution varies greatly between clients.
For this reason, methods such as PruneFL exhibit mask
instability problems in the same setting.
FedDST uses magnitudes of weights and ``votes'' from clients
to solve this problem.

\paragraph{Accuracy Gains from Dynamic Sparsity: A Spatial-Temporal Ensembling Effect.}
Our experiments show that FedDST gains not only communication/computational savings, but also performance improvements: this is particularly notable in highly non i.i.d. settings. We attribute the ``less is more" phenomenon to an underlying spatial-temporal ``ensembling effect" allowed uniquely by FedDST.

\underline{On the ``spatial" side}, we refer to the fact that in FedDST, one mask is sent from the cloud to all clients, yet each client can re-adjust their mask and re-sample the weights according to its non i.i.d. local data. These new sparse masks and weights are periodically re-assembled in the cloud, reminiscent of the famous model sub-sampling and ensembling effect of ``dropout" \cite{hinton2012improving}, now along FL's spatial dimension (across clients). That is, each client can be viewed as a differently sampled subnetwork of the dense cloud model (not random, but ``learned dropout"), and such can provide a regularization effect on training more robust weights for the cloud model by ensembling those subnetwork weights. Note that this similar effect does not take place in PruneFL, where all clients share one mask at each time. While LotteryFL also allows each client to have its own mask, it comes with heavy local computational overhead. \underline{On the ``temporal" side}, FedDST explores the mask space through time while also learning weights $\theta^r$. Taken together, it allows for a continuous parameter exploration across training, taming a \textit{space-time over-parameterization} \cite{intimeoverparameterization}, which can significantly improve the expressibility and generalizability of sparse training.

\paragraph{Extending FedDST to Other FL Frameworks} From the basic example above, FedDST can easily accommodate different local and global optimizers, as described in \cite{reddi2021adaptive}. Algorithm 1 shows a general outline of FedDST, where local training can use
any optimizer as needed. For example, we use SGD with momentum
as described in \cite{fedavgm} as the local optimizer in our
experiments. The pseudo-gradient generated by the
aggregated sparse updates can also be used with other
global optimizers.

FedDST is also compatible with FedProx \cite{fedprox}
and its proximal penalty $\norm{\theta_c^r - \theta^r}_2$.
However, if this penalty is directly used for mask readjustment,
the penalty will act as a weight decay term on weights that have been
pruned out. These weights will thus be less likely to be selected for
regrowth. Therefore, in client growth, we use gradients
corresponding to the loss without the proximal term.

\subsection{Communication and Local Training Savings} \label{analysis}
\textbf{Communication Analysis:} FedDST provides significant competitive bandwidth savings at both uploading and downloading links.
Let $n$ denote the number of parameters in the network. In FedDST,
the same sparsity $S$ is maintained at each round and masks are only
uploaded and downloaded a maximum of once every $\Delta R$ rounds,
so FedDST has an average upload and download cost of
$\left(32(1-S) + \frac{1}{\Delta R}\right)n$ bits per client per round
before $R_\mathit{end}$, and a cost of $32(1-S)n$ after $R_\mathit{end}$.
Furthermore, the maximum upload or download cost at any client is
$\left(32(1-S) + 1\right)n$ bits, so FedDST successfully avoids placing
the burden of large uploads on any one client for any round.
We believe that these upload cost savings should also help
significantly to combat the straggler problem \cite{fedprox, problems}
in practice, as clients with much slower connections will
never be forced to upload full models or gradients.
Furthermore, in commercial FL systems with a large number
of clients, it is likely that a particular client will only be
selected once. With this upper limit on bandwidth costs, FedDST
also makes it practical to learn models via FL, even on
cellular networks.

This is in contrast to PruneFL, which requires clients to send
full gradients to the server every $\Delta R$ rounds,
leading to an average upload cost of
$\left(32(1-S) + \frac{32}{\Delta R}\right)n$ bits per client per round,
and a maximum upload cost of $\left(32(1-S) + 32\right)n$.
Thus FedDST is no worse than PruneFL during normal rounds, and
at sparsity $S=0.8$, FedDST is $5\times$ cheaper than PruneFL per round
with respect to upload cost during mask readjustment
rounds. Finally, PruneFL relies on a single client to provide an
initial sparsity pattern. Because of local client
heterogeneity in FL, the mask produced by one client tends to reflect
the local data distribution of that client. For the same reason, later
updates to the sparsity pattern exhibit instability.
FedDST completely bypasses these problems 
by starting with a random mask followed by the sparse weighted average
aggregation.

\textbf{Computation Analysis:}
FedDST also saves considerable local computational workloads in FL by maintaining
sparse networks throughout the FL process. No part of FedDST requires dense training. In terms of FLOP savings, this allows us to skip
most of the FLOPs in both training and inference, proportional to the
sparsity of the model.
For example, at $80\%$ sparsity, for forward computation on the network we use on CIFAR-10,
only $0.8$ MFLOPs are required, while $4.6$ MFLOPs are required for the dense network.
For the same reason, the experiments against cumulative upload data caps
also roughly reflect the accuracy at different FLOP limits.

Note that following the convention of \cite{Mocanu2018,rigl,intimeoverparameterization}, FedDST so far only considers element-wise unstructured sparsity. Unstructured sparsity was traditionally considered less translatable into real hardware benefits due to irregular access \cite{wen2016learning}. However, at 70\%-90\% unstructured sparsity, XNNPack \cite{elsen2020fast} recently showed significant speedups over dense baselines on smartphones, motivating our future work to optimize practical local training speedups on a FL hardware platform, and to incorporate structured sparsity in dynamic sparse training \cite{earlybird}.

\section{Experiments}  \label{experiments}
\begin{table}[!t]
\centering
\caption{Accuracy of FedDST and other methods given cumulative
upload bandwidth limits, on non-iid MNIST.
We fix $S=0.8$ for sparse methods, $\alpha=0.05$ for DST methods, and
$\mu=1$ for the proximal penalty.}
\begin{tabular}{*5c}
\toprule

& \multicolumn{4}{c}{Best accuracy encountered at} \\
& \multicolumn{4}{c}{cumulative upload capacity [GiB]} \\
Method & 1 & 2 & 3 & 4 \\
\midrule
FedAvgM & 85.25&96.32&97.16&97.53 \\ % fedavg20.log
FedProx ($\mu=1$) & 82.34&95.84&97.16&97.54\\
FedAvgM bfloat16 & 77.41&90.13&96.88&97.46 \\
RandomMask %($S=0.8$)
& 93.61&96.89&97.5&97.72 \\
GraSP %($S=0.8$)
& 61.95&86.06&94.15&96.29 \\
PruneFL & 78.12&89.29&91.65&93.26 \\
FedDST %($S=0.8, \alpha=0.05$)
& \textbf{96.10}&\textbf{97.35}&\textbf{97.67}&\textbf{97.83} \\
FedDST+FedProx %($S=0.8, \alpha=0.05, \mu=1$)
& 95.35&96.97&97.26&97.81 \\
\bottomrule
\end{tabular}
\label{table:mnist}
\end{table}

\begin{table}[!t]
\centering
\caption{Accuracy of FedDST and other methods given cumulative
upload bandwidth limits, on non-iid CIFAR-10. We fix $S=0.8$ for sparse methods, $\alpha=0.001$ for FedDST alone,
$\alpha=0.01$ for FedDST with the proximal penalty, and
$\mu=1$ for the proximal penalty.}
\begin{tabular}{*5c}
\toprule

& \multicolumn{4}{c}{Best accuracy encountered at} \\
& \multicolumn{4}{c}{cumulative upload capacity [GiB]} \\
Method & 4 & 8 & 12 & 16 \\
\midrule
FedAvgM & 24.43&33.87&37.07&40.52 \\ % cifar10_fedavg20.log
FedProx ($\mu=1$) & 23.54&34.01&39.08&42.56 \\
FedAvgM bfloat16 & 22.58&34.05&37.10&41.65 \\
RandomMask %($S=0.8$)
& 33.98&41.86&45.99&48.01 \\
GraSP %($S=0.8$) 
& 15.68&29.5&39.7&44.85 \\
PruneFL & 17.37&25.3&30.88&35.29\\
FedDST %($S=0.8, \alpha=0.001$)
& \textbf{35.41}&42.27&\textbf{46.72}&\textbf{50.67} \\
FedDST+FedProx %($S=0.8, \alpha=0.01, \mu=1$)
& 33.03&\textbf{43.18}&46.66&49.69 \\
\bottomrule
\end{tabular}
\label{table:cifar10}
\end{table}

\begin{table}[!t]
\centering
\caption{Accuracy of FedDST and other methods given cumulative
upload bandwidth limits, on non-iid CIFAR-100. We fix $S=0.5$ for sparse methods, $\alpha=0.01$ for FedDST, and $\mu=1$ for the proximal penalty.}
\begin{tabular}{*5c}
\toprule

& \multicolumn{4}{c}{Best accuracy encountered at} \\
& \multicolumn{4}{c}{cumulative upload capacity [GiB]} \\
Method & 8 & 16 & 24 & 32 \\
\midrule
FedAvgM & 6.66&9.29&10.13&10.94 \\ % cifar10_fedavg20.log
FedProx ($\mu=1$) & 2.74&3.87&4.42&5.12 \\
FedAvgM bfloat16 & 7.78 & 9.92 & 10.92 & 12.02 \\
RandomMask %($S=0.8$)
& 7.15 & 8.65 & 9.41 & 9.69 \\
GraSP %($S=0.8$) 
& 4.45 & 6.61 & 7.78 & 8.37 \\
PruneFL & 5.78 & 8.10 & 9.44 & 10.02\\
FedDST %($S=0.8, \alpha=0.001$)
& 9.14&11.30&13.18&13.96 \\
FedDST+FedProx %($S=0.8, \alpha=0.01, \mu=1$)
& \textbf{9.40} & \textbf{11.29} & \textbf{13.46} & \textbf{14.57} \\
\bottomrule
\end{tabular}
\label{table:cifar100}
\end{table}

We use MNIST \cite{lecun2010mnist} and CIFAR-10 \cite{Krizhevsky09learningmultiple}
datasets distributed among clients in a
``pathologically non-iid'' setting, similar to~\cite{fedavg}
and matching the datasets used in~\cite{lotteryfl}.
We assume a total pool of $400$ clients. Each client
is assigned $2$ classes and given $20$ training images from each class.
To distribute CIFAR-100 \cite{Krizhevsky09learningmultiple} in a non-iid fashion,
we use a $\mathrm{Dirichlet}(0.1)$ distribution for each class
to distribute its samples among $400$ clients, as in \cite{fedma, moon, fedimbalance}.
These distributions represent a challenging and realistic training environment, with
only a fraction of the training data available and
distributed in a severely non-iid fashion among clients.
We provide more details for all datasets in the appendix.

\begin{figure}[!t]
\begin{subfigure}[b]{0.45\textwidth}
\centering
\includegraphics[scale=0.5]{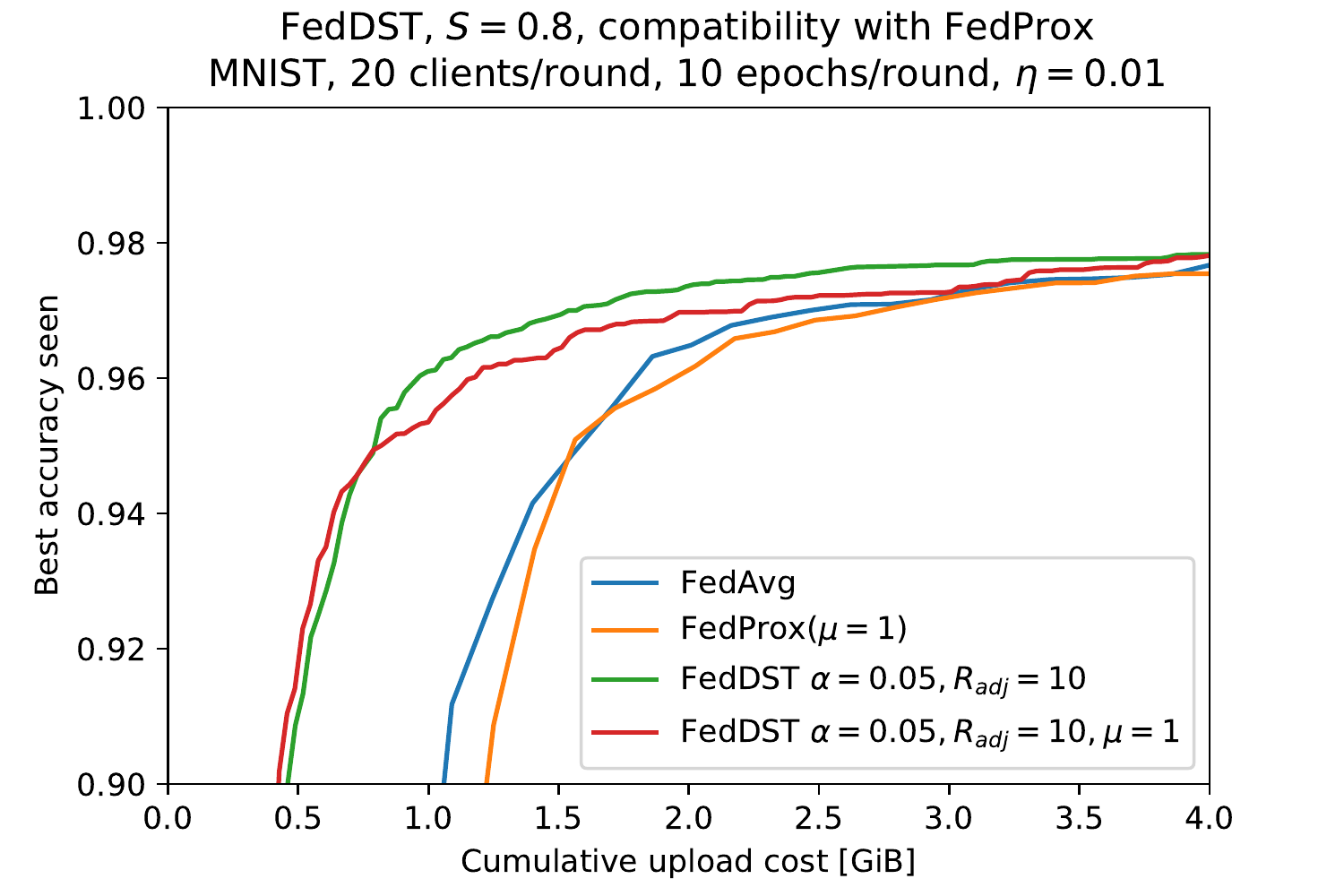}
\caption{FedDST+FedProx on non-iid MNIST}
\label{fig:feddst_mnist_compat}
\end{subfigure}
\begin{subfigure}[b]{0.45\textwidth}
\centering
\includegraphics[scale=0.5]{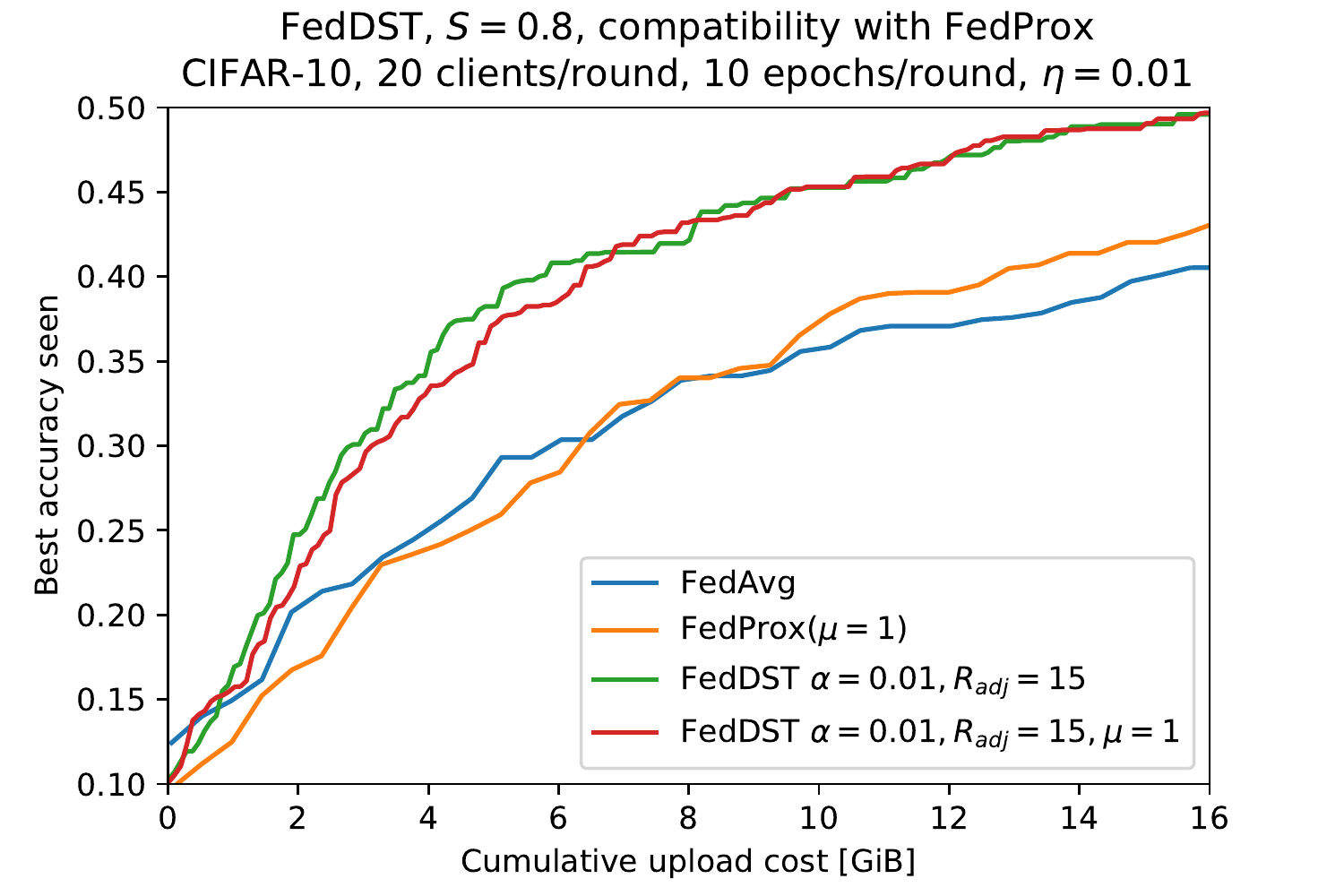}
\caption{FedDST+FedProx on non-iid CIFAR-10}
\label{fig:feddst_cifar10_compat}
\end{subfigure}
\begin{subfigure}[b]{0.45\textwidth}
\centering
\includegraphics[scale=0.5]{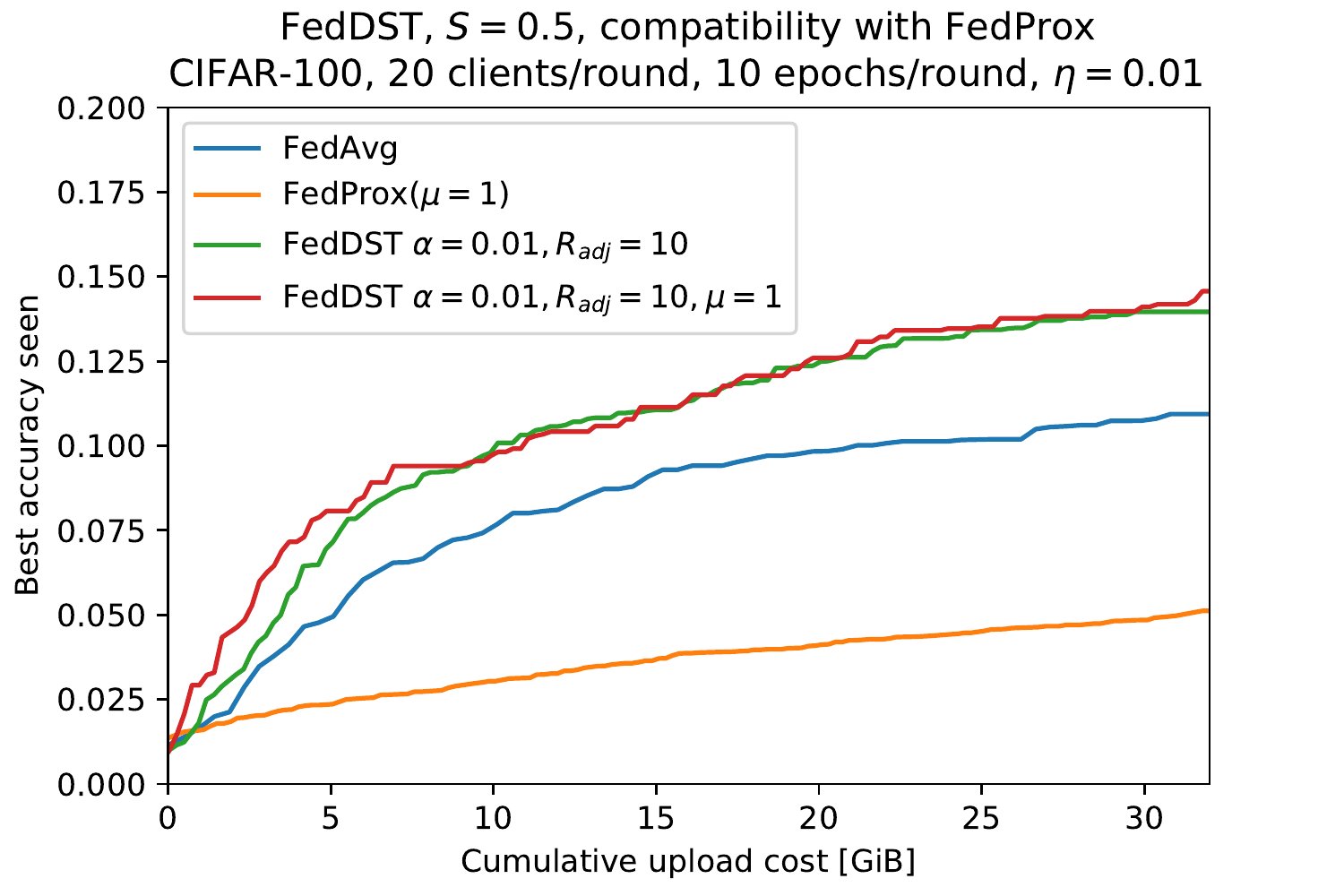}
\caption{FedDST+FedProx on non-iid CIFAR-100}
\label{fig:feddst_cifar100_compat}
\end{subfigure}

\caption{FedDST is compatible with other popular FL frameworks, such as FedProx. This shows that FedDST is a general framework that can also apply to other optimizers.
}
\label{fig:feddst_compat}
\end{figure}

\subsection{Main Results}
We compare FedDST to competitive state-of-the-art
baselines. Both FedAvgM~\cite{fedavgm}
and FedProx~\cite{fedprox} are designed for non-iid settings.
We include PruneFL~\cite{prunefl} as the other
method involving dynamic sparsity in FL. As mentioned before,
PruneFL relies on stochastic gradients uploaded by clients,
so its masks exhibit instability between reconfigurations.
In the cases where PruneFL converged, it selected
an all-ones mask, recovering plain FedAvg.
Note that because FedDST aims to produce a single model in each round
that performs well on all clients, we do not compare to
algorithms producing separate models for each client, such as
LG-FedAvg \cite{lg_fedavg} and LotteryFL \cite{lotteryfl}.

For RandomMask, we randomly sample weights at the server,
then perform layer-wise magnitude pruning, following the
ERK sparsity distribution \cite{rigl}, before the first round,
and perform FedAvgM on this sparse network. The
random mask selected is held constant as the global and local sparsity
mask throughout training. RandomMask represents a strong communication-efficient
baseline for this non-iid setting, in which the noisy
magnitudes of stochastic gradients cause methods
that rely on gradient aggregation, such as PruneFL, to fail.
Furthermore, the data heterogeneity in this environment causes
GraSP to select a mask that works well for the initially chosen client
but does not work well as a global mask.
Despite this challenging environment, FedDST still produces significant
accuracy improvements over the strong RandomMask baseline.

Tables \ref{table:mnist}, \ref{table:cifar10}, and \ref{table:cifar100} provide accuracies
achieved by FedDST and other algorithms given certain upload bandwidth limits.
We report the best test accuracy seen before each limit, averaged across 10 runs.
FedDST consistently provides the best performance at any upload limit, halving
upload cost with respect to FedAvgM in all settings we tested.

\paragraph{Compatibility with FedProx}
Our experimental results also confirm that FedDST is compatible with other
FL frameworks, including FedProx. For this experiment, we add FedProx's
proximal term and adjust FedDST's growth criterion as described in the overview.
Figure \ref{fig:feddst_compat}
shows that applying FedDST to FedProx consistently improves its performance
across datasets. FedDST is therefore a general framework that can be applied
to various FL optimizers.

\begin{figure}[!t]
\begin{subfigure}[b]{0.45\textwidth}
\centering
\includegraphics[scale=0.5]{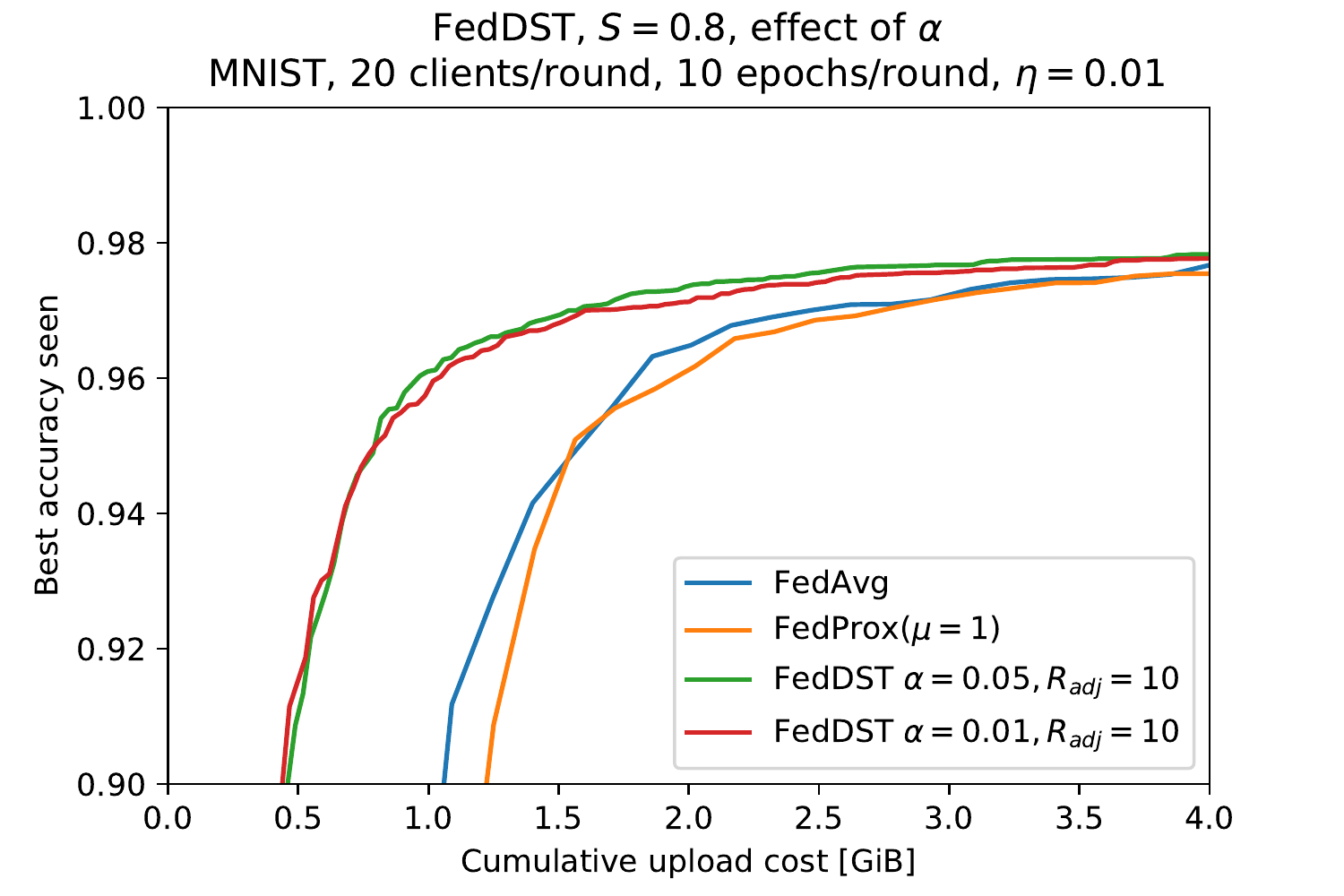}
\caption{FedDST on non-iid MNIST}
\label{fig:feddst_mnist_alpha}
\end{subfigure}
\begin{subfigure}[b]{0.45\textwidth}
\centering
\includegraphics[scale=0.5]{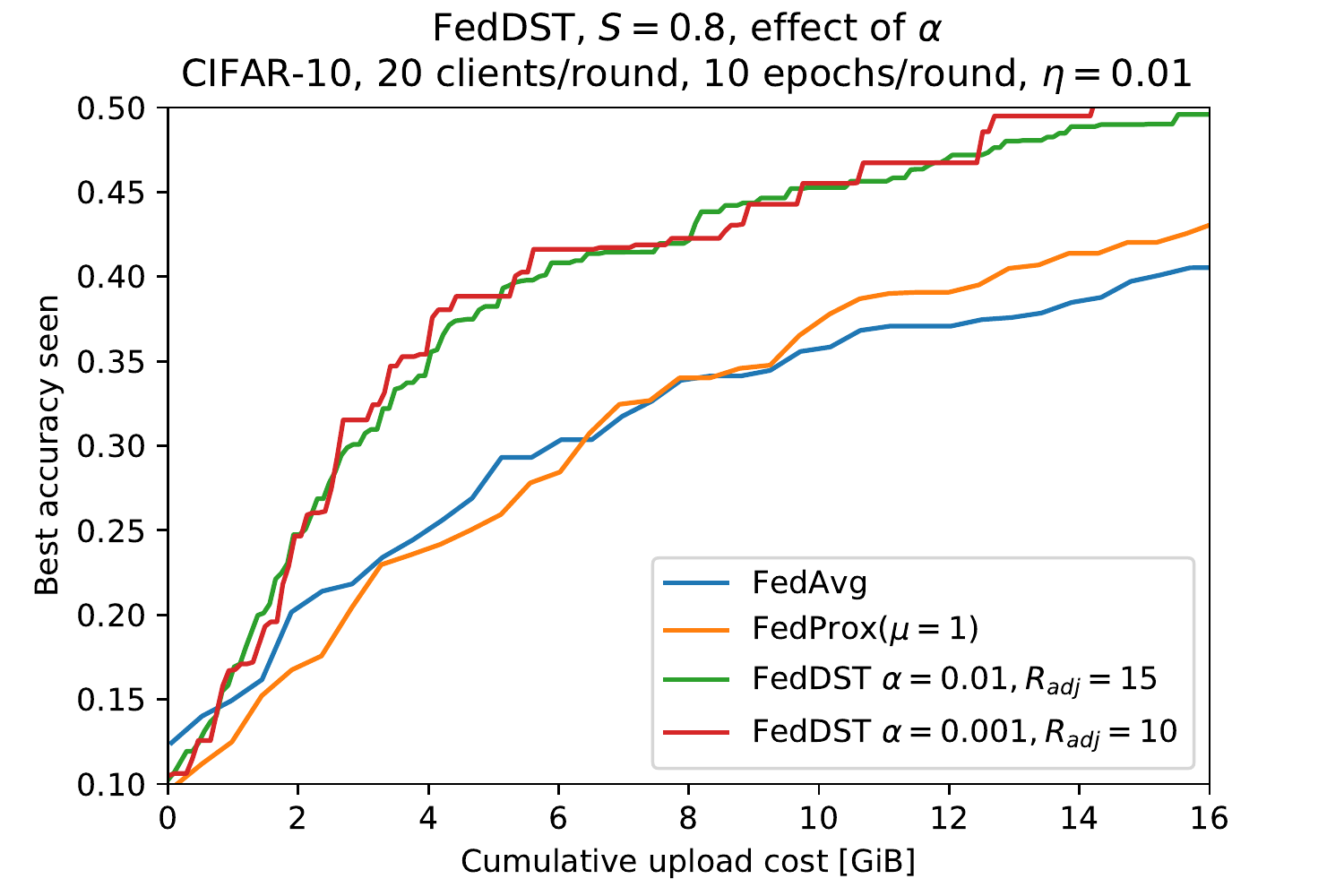}
\caption{FedDST on non-iid CIFAR-10}
\label{fig:feddst_cifar10_alpha}
\end{subfigure}
\caption{FedDST is insensitive to certain variations in $\alpha$.}
\label{fig:feddst_alpha}
\end{figure}

\subsection{Insensitivity to Variations in $\alpha$}
\label{alpha}
In Figure \ref{fig:feddst_alpha}, we show that
FedDST performs well with any reasonable value of $\alpha$,
even in the non-iid setting we explore in this paper.
The non-iid distribution
targeted by FedDST leads to noisy stochastic gradients produced by
clients. Thus, directly using the magnitudes of gradients to produce
mask readjustments causes large variations in the mask.
FedDST effectively sidesteps this problem by only using the top
local stochastic gradients to ``vote'' for particular weight indices.
The parameter $\alpha$ specifies how many of these ``votes''
should be submitted to the server. However, suitable values for
$\alpha$ are still smaller than in RigL; for
$\alpha \in [0.001, 0.05]$, FedDST significantly outperforms other
methods. 

\begin{figure}[!t]
\begin{subfigure}[b]{0.45\textwidth}
\centering
\includegraphics[scale=0.5]{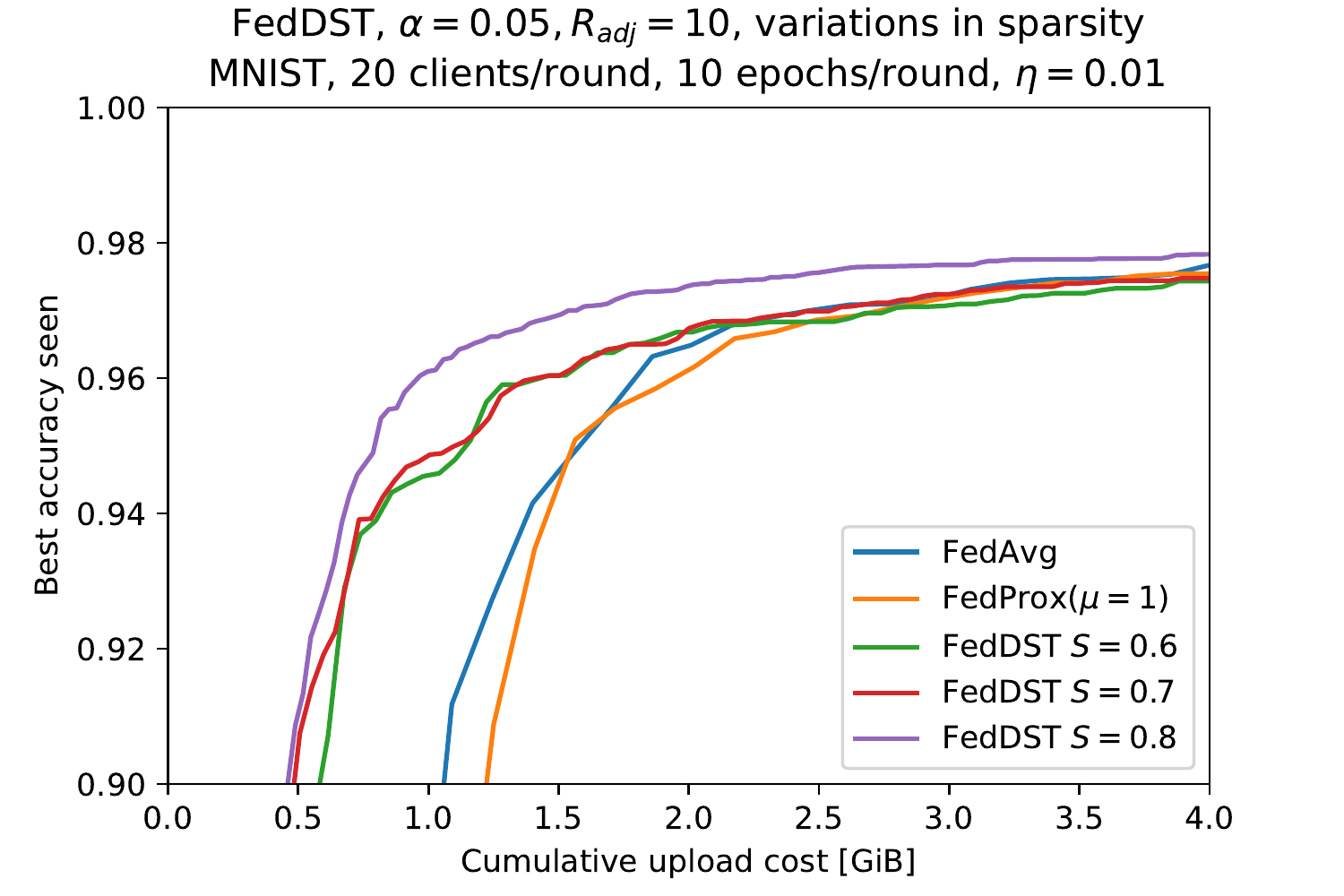}
\caption{FedDST with varying sparsity on non-iid MNIST}
\label{fig:feddst_mnist_sparsity}
\end{subfigure}
\begin{subfigure}[b]{0.45\textwidth}
\centering
\includegraphics[scale=0.5]{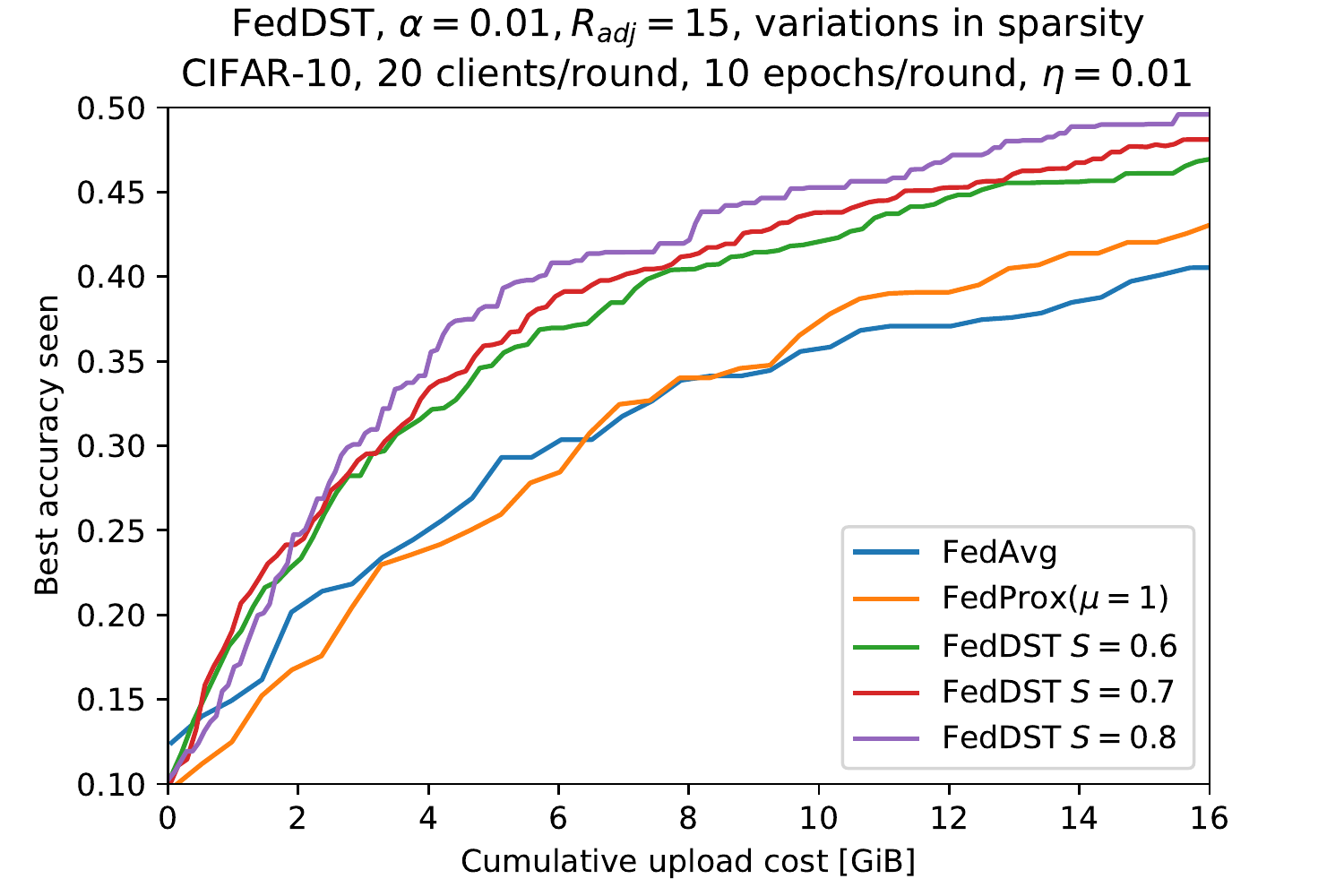}
\caption{FedDST with varying sparsity on non-iid CIFAR-10}
\label{fig:feddst_cifar10_sparsity}
\end{subfigure}
\begin{subfigure}[b]{0.45\textwidth}
\centering
\includegraphics[scale=0.5]{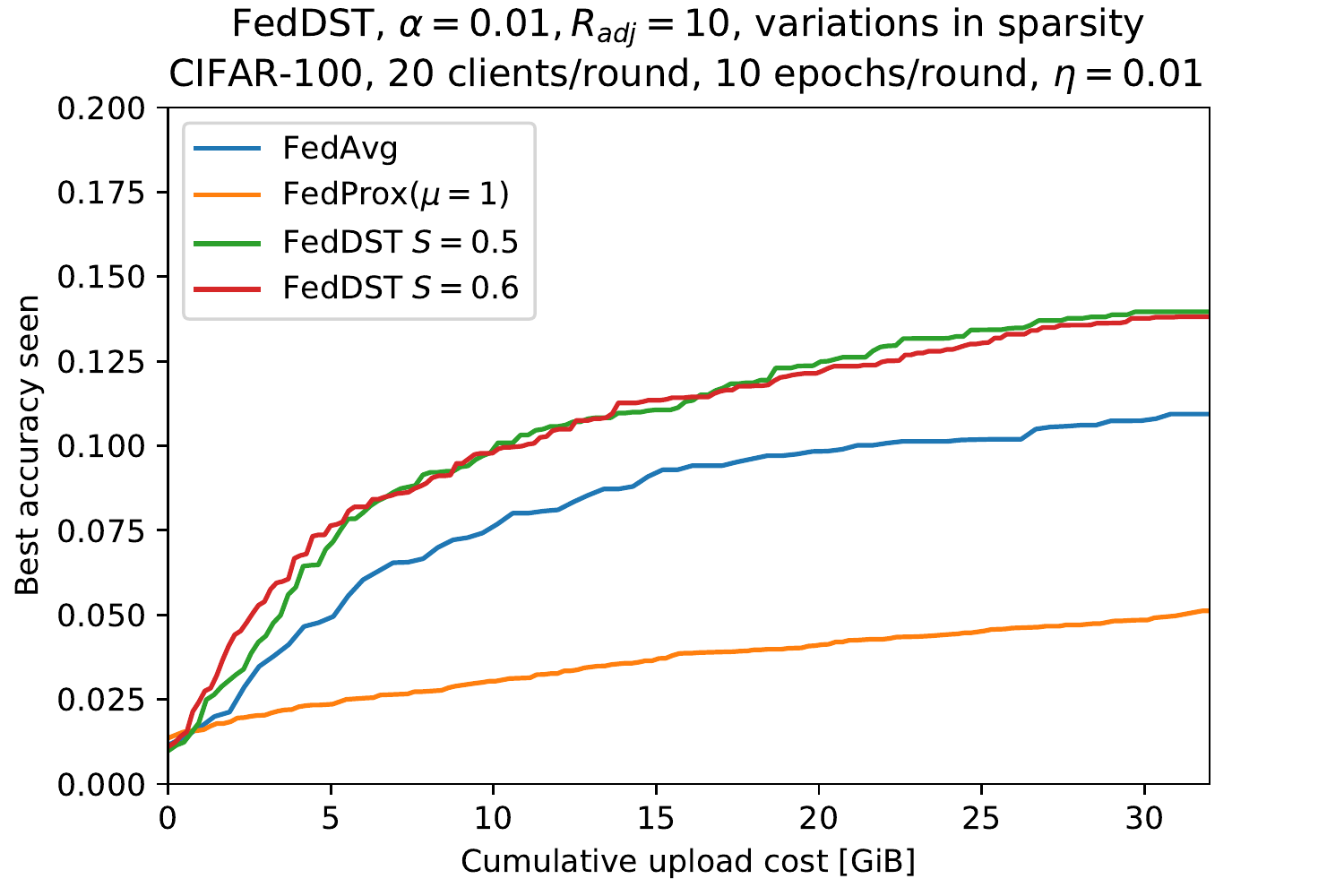}
\caption{FedDST with varying sparsity on non-iid CIFAR-100}
\label{fig:feddst_cifar100_sparsity}
\end{subfigure}
\caption{FedDST is robust to sparsity variations, and performs the best at reasonable sparsities.}
\label{fig:feddst_sparsity}
\end{figure}

\subsection{Performance at Different Sparsity Levels}
In Figure \ref{fig:feddst_sparsity}, we show that FedDST's
performance is robust to variations in sparsity.
As sparsity directly leads to communication savings,
FedDST performs best at relatively high sparsities, even on
the lightweight NNs we test here.
However, even at untuned sparsity levels, FedDST performs
reasonably well and converges much more quickly than FedAvgM,
especially at the beginning of the FL process.
Hence we run all FedDST experiments
at sparsity $0.5$ or $0.8$.

\section{Conclusion and Broader Impacts}
We introduce \textit{Federated Dynamic Sparse Training} (FedDST), a
powerful framework for communication-efficient federated learning
via dynamic sparse training, which works well even on pathologically
non-iid datasets.
We show experimentally that FedDST consistently
outperforms competing algorithms, producing up to $3\%$ better accuracy than
FedAvgM with half the upload bandwidth on non-iid CIFAR-10.
We further demonstrate that FedDST is compatible with other popular
federated optimization frameworks such as FedProx.
Our results indicate a bright future for sparsity
in even the most difficult FL settings.

FL has the potential to enable learning NNs in situations in which
privacy of user data is of paramount importance. For example,
medical datasets are bound by strong legal restrictions and cannot
be exchanged between clients. Furthermore, FL can provide stronger
privacy guarantees in existing machine learning settings involving
user data. With its focus on improving communication efficiency
and performance, FedDST makes more FL settings practical in the wild.
%expanding user privacy in both existing and new ML domains.

\section{Acknowledgements}

% Use \bibliography{yourbibfile} instead or the References section will not appear in your paper

The authors would like to recognize that the corresponding author, Xiaohan Chen,
was the other student author driving the idea and experimental design in this paper.

Portions of this research (by S. Bibikar and H. Vikalo) were
sponsored by the Army Research Office and were accomplished under Cooperative Agreement
Number W911NF-19-2-0333. The views and conclusions contained in this document are
those of the authors and should not be interpreted as representing the official policies, either
expressed or implied, of the Army Research Office or the U.S. Government. The U.S. Government
is authorized to reproduce and distribute reprints for Government purposes notwithstanding any
copyright notation herein.

X. Chen and Z. Wang's research were supported in part by the NSF Real-Time Machine Learning program (Award Number: 2053279), and the NSF AI Institute for Foundations of Machine Learning (IFML).

\bibliography{references}

\end{document}

\appendix

\section{Experimental Details}

For the experiments, we adopt FedAvgM~\cite{fedavgm} as the optimizer,
with a momentum of $0.9$, fixed client learning rate $\eta = 0.01$,
$L_2$ regularization coefficient $0.001$,
and $|C_i| = 20$ clients selected at random each round.
For MNIST, we set $R_\mathit{end}=200$, and for CIFAR-10, we set
$R_\mathit{end}=1000$. The details for the models used \cite{lotteryfl}
are provided in Table \ref{table:models}.

\begin{table}[h]
\centering
\caption{Model details for all experiments reported in this paper.}
\label{table:models}
\begin{tabular}{cc}
\toprule
\textbf{MNIST CNN} & \textbf{CIFAR-10 CNN} \\
\midrule
Conv2d$(1, 10, 5)$ & Conv2d$(3, 6, 5)$ \\
MaxPool $3\times 3$, stride $1$ & MaxPool $3\times 3$, stride $1$ \\
ReLU & ReLU \\
Conv2d$(10, 20, 5)$ & Conv2d$(3, 6, 5)$ \\
MaxPool $3\times 3$, stride $1$ & MaxPool $3\times 3$, stride $1$ \\
ReLU & ReLU \\
\midrule
$50$ fully-connected units & $120$ fully-connected units \\
ReLU & ReLU \\
$10$ fully-connected units & $84$ fully-connected units \\
& ReLU \\
& $10$ fully-connected units \\
\bottomrule
\end{tabular}
\end{table}

%For FedDST+FedProx, we add FedProx's proximal term according to
%section \ref{feddst_overview}.
\paragraph{Baselines.}
For the bfloat16 results, we quantize only the uploads from clients to bfloat16.
Bfloat16 is an aggressive truncation of standard 32-bit floating-point
values accomplished by removing the lower 16 bits of the mantissa.
For %SNIP \cite{lee2018snip} and 
GraSP~\cite{Wang2020Picking}, we 
randomly sample a client before training and perform single-shot
pruning using one minibatch of that client's data, running the code by Tanaka et al\@.~\cite{tanaka2020}.
For PruneFL, we use the MIT-licensed code provided by \cite{prunefl}
to sample per-layer times.
For unstructured pruning, we include a modified version of
\verb|prune.py| from PyTorch.

\paragraph{Compute and Evaluation.}
We report the average validation
accuracy across all clients and the upload cost; the results
are averaged across $8$ runs.
We conducted the experiments on AMD Vega 20 (ROCm)
and Nvidia GTX 1080 Ti (CUDA)
cards; each table row or plotted line requires $12$
GPU hours for MNIST, and $40$ GPU hours for CIFAR-10.
Due to high instabilities exhibited by FL algorithms
in the pathologically non-iid setting, instead of
the accuracy at a particular round, we report the maximum
accuracy attained given a particular upload cost, which we find to more consistently assess each algorithm's capability.

\paragraph{Error.}
In Tables \ref{table:mnist_err} and \ref{table:cifar10_err},
we report standard deviations for the average values reported in Tables \ref{table:mnist} and \ref{table:cifar10}. Note that
individual runs of all the tested algorithms experience large fluctuations in accuracy over the course of FL training. For this reason, we report the maximum accuracy attained.

\begin{table*}[!t]
\centering
\caption{Accuracy of FedDST and other methods given cumulative
upload bandwidth limits, on non-iid MNIST. We report the best test accuracy encountered before
the specified bandwidth limit, averaged across 10 runs.\vspace{0.5em}}

\small
\begin{tabular}{*5c}
\toprule

& \multicolumn{4}{c}{Best accuracy encountered at} \\
& \multicolumn{4}{c}{cumulative upload capacity [GiB]} \\
Method & 1 & 2 & 3 & 4 \\
\midrule
FedAvgM & $85.25 \pm 23.96$&$96.32 \pm 0.47$&$97.16 \pm 0.39$&$97.53 \pm 0.24$ \\ % fedavg20.log
FedProx ($\mu=1$) & $82.34 \pm 27.74$&$95.84 \pm 0.79$&$97.16 \pm 0.2$&$97.54 \pm 0.11$\\
FedAvgM bfloat16 & $77.41 \pm 24.38$&$90.13 \pm 13.18$&$96.88 \pm 0.58$&$97.46 \pm 0.21$ \\
RandomMask ($S=0.8$) & $93.61 \pm 3.8$&$96.89 \pm 0.25$&$97.5 \pm 0.06$&$97.72 \pm 0.05$ \\
%SNIP ($S=0.8$) & 94.38&97.09&97.52&97.77 \\
GraSP ($S=0.8$) & $61.95 \pm 31.64$&$86.06 \pm 17.27$&$94.15 \pm 6.56$&$96.29 \pm 2.59$ \\
PruneFL & $78.12 \pm 17.25$&$89.29 \pm 7.65$&$91.65 \pm 5.6$&$93.26 \pm 5.15$ \\
%Stochastic Sign-SGD & TODO & TODO & TODO & TODO\\
%FedDST ($S=0.8, \alpha=0.05$, visc.) & 95.46&96.93&97.34&97.54 \\
FedDST ($S=0.8, \alpha=0.05$) & $\mathbf{96.1 \pm 0.37}$&$\mathbf{97.35 \pm 0.28}$&$\mathbf{97.67 \pm 0.2}$&$\mathbf{97.83 \pm 0.17}$ \\
FedDST+FedProx & $95.35 \pm 0.68$&$96.97 \pm 0.25$&$97.26 \pm 0.13$&$97.81 \pm 0.12$ \\
 ($S=0.8, \alpha=0.05, \mu=1$) & \\
\bottomrule
\end{tabular}
\label{table:mnist_err}
\end{table*}

\begin{table*}[!t]
\vspace{-1em}
\centering
\caption{Accuracy of FedDST and other methods given cumulative
upload bandwidth limits, on non-iid CIFAR-10. We report the best test accuracy encountered before
the specified bandwidth limit, averaged across 10 runs.\vspace{1em}}
\small
\begin{tabular}{*5c}
\toprule

& \multicolumn{4}{c}{Best accuracy encountered at} \\
& \multicolumn{4}{c}{cumulative upload capacity [GiB]} \\
Method & 4 & 8 & 12 & 16 \\
\midrule
FedAvgM & $24.43 \pm 3.15$&$33.87 \pm 2.12$&$37.07 \pm 1.63$&$40.52 \pm 1.07$ \\ % cifar10_fedavg20.log
FedProx ($\mu=1$) & $23.54 \pm 0.71$&$34.01 \pm 3.41$&$39.08 \pm 1.61$&$42.56 \pm 3.23$ \\
FedAvgM bfloat16 & $22.58 \pm 3.98$&$34.05 \pm 2.36$&$37.1 \pm 2.63$&$41.65 \pm 2.24$ \\
RandomMask ($S=0.8$) & $33.98 \pm 4.06$&$41.86 \pm 3.64$&$45.99 \pm 2.95$&$48.01 \pm 2.61$ \\
%SNIP ($S=0.8$) & TODO & TODO & TODO & TODO \\
GraSP ($S=0.8$) & $15.68 \pm 6.34$&$29.5 \pm 9.12$&$39.7 \pm 5.34$&$44.85 \pm 2.67$ \\
PruneFL & $17.37 \pm 6.58$&$25.3 \pm 7.48$&$30.88 \pm 4.3$&$35.29 \pm 5.93$\\
%Stochastic Sign-SGD & TODO & TODO & TODO & TODO\\
%FedDST ($S=0.8, \alpha=0.001$, visc.) & 34.9&42.22&45.19&46.79 \\
FedDST ($S=0.8, \alpha=0.001$) & $\mathbf{35.41 \pm 1.1}$&$42.27 \pm 0.52$&$\mathbf{46.72 \pm 2.03}$&$\mathbf{50.67 \pm 2.38}$ \\
FedDST+FedProx  & $33.03 \pm 1.02$&$\mathbf{43.18 \pm 1.26}$&$46.66 \pm 1.15$&$49.69 \pm 1.26$ \\
($S=0.8, \alpha=0.01, \mu=1$) & \\
\bottomrule
\end{tabular}
\vspace{-0.5em}
\label{table:cifar10_err}
\end{table*}

\section{Using the FedDST Experiment Code}

\paragraph{Dependencies.}  The attached code requires
\verb|Python|, version 3.6 or greater,
\verb|PyTorch|, \verb|torchvision|, and \verb|tqdm|.
The dataset code can be fetched by running
\verb|git submodule init|, followed by
\verb|git submodule update|,
in the directory containing the unpacked source code.

\paragraph{Sample Command Lines.}
We briefly provide example command lines, usable with the
attached experiment code, that will execute
selected experiments, in Table \ref{table:command}.

\begin{table*}[!t]
\centering
\caption{Sample command lines for the FedDST experiment code.}
\label{table:command}
\begin{tabular}{cl}
\toprule
\textbf{Experiment} & \textbf{Command Line} \\
\midrule
FedAvg on CIFAR-10 & \verb|python3 dst.py --dataset cifar10| \\
& \verb|--sparsity 0.0| \\
\midrule
FedProx on CIFAR-10 & \verb|python3 dst.py --dataset cifar10| \\
$(\mu=1)$& \verb|--sparsity 0.0 --prox 1| \\
\midrule
FedDST on CIFAR-10 &
\verb|python3 dst.py --dataset cifar10| \\
$(S=0.8, \alpha=0.01, R_\mathit{adj}=15)$ & \verb|--sparsity 0.8 --readjustment-ratio 0.01| \\
& \verb|--rounds-between-readjustments 15| \\
\midrule
FedDST+FedProx on CIFAR-10 &
\verb|python3 dst.py --dataset cifar10| \\
$(S=0.8, \alpha=0.01, R_\mathit{adj}=15, \mu=1)$ & \verb|--sparsity 0.8 --readjustment-ratio 0.01| \\
& \verb|--rounds-between-readjustments 15 --prox 1| \\
\midrule
RandomMask on MNIST &
\verb|python3 dst.py --dataset mnist| \\
$(S=0.8)$ & \verb|--sparsity 0.8 --readjustment-ratio 0.0| \\
\midrule
PruneFL on MNIST &
\verb|python3 prunefl.py --dataset mnist| \\
& \verb|--rounds-between-readjustments 50| \\
& \verb|--initial-rounds 1000| \\
\bottomrule
\end{tabular}
\end{table*}

% Use \bibliography{yourbibfile} instead or the References section will not appear in your paper
\bibliography{references}